\documentclass[conference]{IEEEtran}
\IEEEoverridecommandlockouts
\usepackage{cite}
\usepackage{amsmath,amssymb,amsfonts}
\usepackage{algorithmic}
\usepackage{graphicx}
\usepackage{textcomp}
\usepackage{lmodern}
\usepackage{mathtools, nccmath}

\usepackage{xcolor}
\usepackage{amsmath,amssymb,amsthm}
\newtheorem{hyp}{Hypothesis}

\usepackage{tabularx,ragged2e,booktabs,caption}
\usepackage{multicol,tabularx,capt-of}
\usepackage{multirow}
\usepackage{lettrine}
\usepackage[export]{adjustbox}
\usepackage{xcolor,colortbl}

\definecolor{Gray}{gray}{0.85}
\definecolor{LightCyan}{rgb}{0.88,1,1}

\newcolumntype{a}{>{\columncolor{Gray}}c}
\newcolumntype{b}{>{\columncolor{white}}c}
\def\BibTeX{{\rm B\kern-.05em{\sc i\kern-.025em b}\kern-.08em
    T\kern-.1667em\lower.7ex\hbox{E}\kern-.125emX}}
\begin{document}

\title{Machine Learning Based Prediction of Future Stress Events in a Driving Scenario}

\author{\IEEEauthorblockN{Joseph Clark, Rajdeep Kumar Nath, and Himanshu Thapliyal}
\IEEEauthorblockA{\textit{Department of Electrical and Computer Engineering, University of Kentucky, Lexington, KY, USA} \\
%\IEEEauthorblockA{\textit{$^{2}$The Maya Angelou Center for Health Equity,  Wake Forest School of Medicine, Winston-Salem, NC, USA} \\
Corresponding Author: hthapliyal@ieee.org}
}

\maketitle
\begin{abstract}
This paper presents a model for predicting a driver's stress level up to one minute in advance. Successfully predicting future stress would allow stress mitigation to begin before the subject becomes stressed, reducing or possibly avoiding the performance penalties of stress. The proposed model takes features extracted from Galvanic Skin Response (GSR) signals on the foot and hand and Respiration and Electrocardiogram (ECG) signals from the chest of the driver. The data used to train the model was retrieved from an existing database and then processed to create statistical and frequency features. A total of 42 features were extracted from the data and then expanded into a total of 252 features by grouping the data and taking six statistical measurements of each group for each feature. A Random Forest Classifier was trained and evaluated using a leave-one-subject-out testing approach. The model achieved 94\% average accuracy on the test data. Results indicate that the model performs well and could be used as part of a vehicle stress prevention system.
\end{abstract}


\section{Introduction}
High levels of stress impair task performance and can lead to accidents while driving \cite{drive_stress}. Stress detection can aid the mitigation of stress only after the subject has become stressed. A method of predicting stress in advance could improve stress mitigation strategies by allowing stress mitigation to begin before the subject enters a stressed state.

Detection of physiological stress has been addressed in numerous papers, either by simple correlation with physiological data\cite{drive_stress2} or by the use of machine learning algorithms \cite{youngjun_stress_detect} \cite{martino_stress_detect}. We have been unable to find any paper which has attempted to use machine learning to predict the stress level of a subject sometime before the subject enters the "stressed" state. Early prediction of stress has the advantage of allowing stress mitigation methods to begin before the subject is stressed, ideally negating the decreased task performance associated with stress.

\begin{figure}[ht]
\centering
\includegraphics[trim= 0cm 0cm 0cm 0cm, scale=0.23]{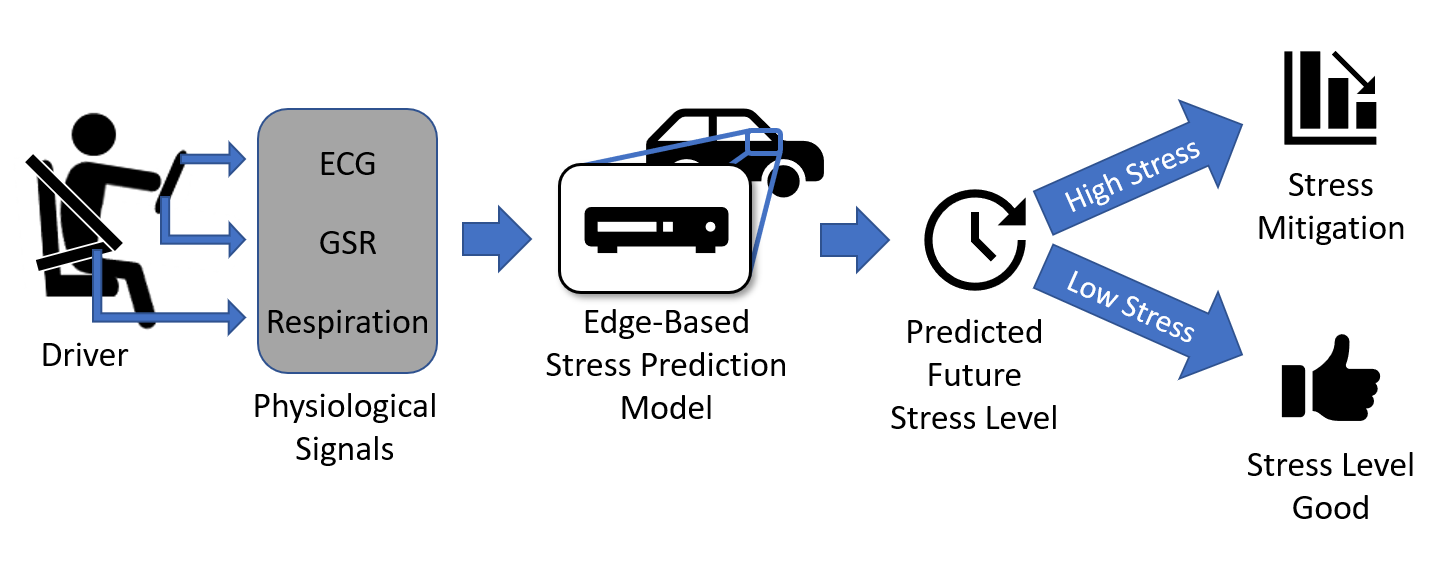}
\caption{Driver Stress Prevention System Overview}
\label{sys_ovr}
\end{figure}

In this work, we propose a model which predicts approaching stress, rather than the current stress level of the subject. If used in conjunction with a car entertainment system or smartphone, this model could be used to launch an intervention to decrease the subject's stress before it rises above a certain threshold. An overview of such a system is presented in Figure \ref{sys_ovr}.

In the stress prevention system, the vehicle has GSR, ECG, and Respiration sensors built-in \cite{stress_sensors} (or otherwise attached, e.g. via a smartwatch worn by the driver). The physiological signals collected by the sensors are then transmitted to an onboard edge computing device which runs the stress prediction model. If the predicted stress level is high, the system can automatically begin personalized stress mitigation strategies \cite{stress_intervention}. The chosen stress mitigation strategy can range from playing music to more advanced strategies such as reducing driver task load by reducing the flow of non-essential information to the driver \cite{healey_stress_detect}. The system would also incorporate driver feedback to improve model accuracy with long-term use.

This paper is organized as follows: Section \ref{prop_work} describes the proposed model and the data set it will be trained on. Section \ref{stress_model} presents the data processing procedure and feature extraction. Section \ref{res} discusses the results, Section \ref{discuss} discusses limitations of the model, and Section \ref{future_research} discusses future research directions for this model. Section \ref{conc} concludes the article.

\section{Proposed Work}
\label{prop_work}

\begin{figure*}[t]
\centering
\includegraphics[trim= 0cm 0cm 0cm 0cm, scale=0.38]{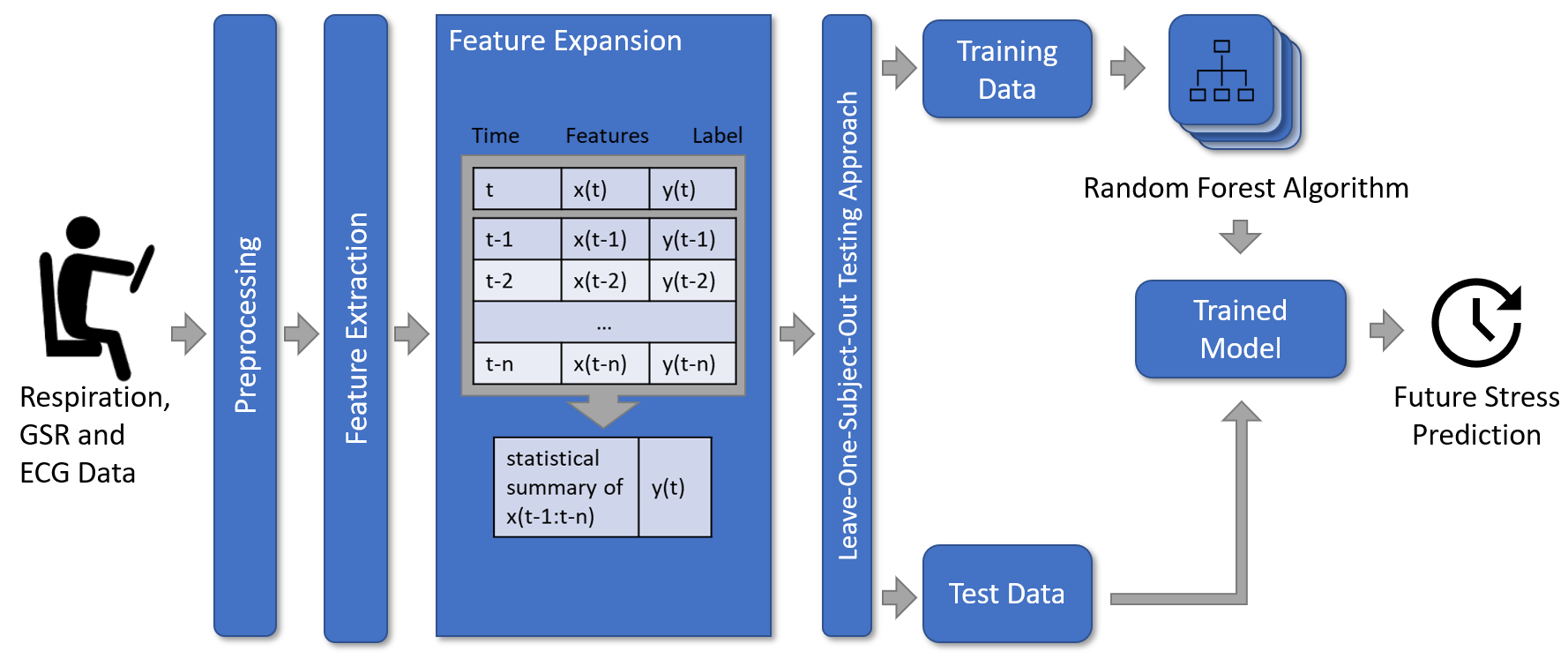}
\caption{Stress Prediction Model Overview}
\label{ovr}
\end{figure*}

Figure \ref{ovr} shows a diagram representing the proposed stress prediction model. The model begins with data preprocessing, feature extraction, and feature expansion. The extracted features are then separated into training and test data following a leave-one-subject-out (LOSO) testing approach. The training data is used to train a random forest classifier, the performance of which is then measured using the test data. We propose the following hypothesis to explore the possibility of predicting stress using this model.

\begin{hyp}
If $x(t)$ denotes the physiological feature space at time $t$ and $y(t)$ denotes the subject's stress level at time $t$, then it is possible to develop a stress model $M$ trained on $x(t-1:t-n)$, where $n$ is the number of time steps used to train the model, that can predict $y(t)$ with good accuracy.
\end{hyp}

We will present the data used to train the stress prediction model, the model itself, and discuss the performance of the model on test data. Four different values of $n$, $n =$ 2, 3, 4, and 5, will be tested in the feature expansion phase of the model and evaluated. We will be training the model with data consisting of six of our seven subjects and we will test the accuracy of the trained model using the remaining subject, following the LOSO testing approach. The experimental data used in this work is a subset of the data collected in \cite{healey_stress_detect}.

\subsection{Data Extraction}

The data set used in this analysis consists of physiological data from 17 subjects as they completed an approximately 20-mile driving route through Boston, obtained from PHYSIONET \cite{healey_stress_detect}.
While there is data from a total of 17 subjects in the data set, only ten of these records (drive 05, 06, 07, 08, 09, 10, 11, 12, 15, and 16) are complete. The remaining records are either missing some of the sensor data or data which differentiates different driving periods. Of the ten records that have complete data, an additional three records (drive 08, 09, and 16) were found to be unsuitable for use in developing this model. Specifically, these three records have gaps in recorded data and data consistency problems.
Each of these records is broken into three distinct driving situations: Rest, City, and Highway. The driving route and experimental protocol were structured such that each drive begins and ends with a Rest period, followed by alternating City and Highway driving periods. This data set is unfortunately also missing information concerning the time that each of these periods begins and ends for each record. This information is also absent from \cite{healey_stress_detect} but is available in \cite{db_times}.

\section{Data Analysis}
\label{stress_model}
The data analysis of the stress prediction model is divided into three phases: (i) Data Preprocessing, (ii) Feature Extraction, (iii) Feature Expansion. Feature selection was considered but did not show any performance improvement. The overview of the system is shown in Figure \ref{ovr}.

\subsection{Data Preprocessing}
The data preprocessing module consists of three stages. The first stage normalizes the GSR, Respiration, and ECG signals to be within the range of 0 to 1. The second stage uses a Butterworth filter of 5 order to filter off signal components that are higher than 1 Hz for the GSR signals, 10 Hz for the respiration signal, and 40 Hz for the ECG signal. The third stage integrates the signals from different time periods of the experiment for feature extraction.

\subsection{Feature Extraction}
The feature extraction module consists of three parts, one for each type of physiological signal. All of the features extracted from the physiological signals were calculated over a running 100 second window with a 50 second overlap. 

\subsubsection{GSR Feature Extraction} A total of 7 features were extracted from both the hand and foot GSR signals. Two features, the mean and variance, were taken from the original signal, while the remainder are based on the peaks in the signal. The peaks are detected by applying a peak finding algorithm \cite{gsr_code} to the first derivative of the GSR signal. %consider figure here
%\begin{figure}[ht]
%\centering
%\includegraphics[trim= 0cm 0cm 0cm 0cm, scale=0.43]{gsr_peak.png}
%\caption{Original GSR signal (top), first derivative of GSR signal with peaks (bottom)}
%\label{gsr_f}
%\end{figure}
The features derived from the peaks are the number of peaks occurring in a window, the sum of the peak amplitudes and duration, and the mean and variance of the peak prominence.
\subsubsection{Respiration Feature Extraction} A total of 6 features were extracted from the respiration signal. There are two statistical features: the mean and variance of the signal, and four frequency features: the power in the 0-0.1 Hz, 0.1-0.2 Hz, 0.2-0.3 Hz, and 0.3-0.4 Hz bands. % consider revising to something like "The chosen bands are each 0.1 Hz wide and range from 0 to 0.4 Hz..."
These were extracted by computing a periodogram for the signal and using and algorithm \cite{resp_code} to calculate the band power on the desired bands.

%Systolic peaks were obtained by using the find$\_peaks$ algorithm. Peaks with height thresholds of 0.4 and widths of 10\% of the sampling rate were classified as systolic peaks. Also, the minimum distance between peaks was set as about 65\% of the sampling rate. Figure \ref{ppg_f} shows the peaks classified as systolic peaks marked by blue cross along with the original PPG signal. Subsequently, these peaks were used to calculate HR and HRV related features
%\begin{figure}[ht]
%\centering
%\includegraphics[trim= 0cm 0cm 0cm 0cm, scale=0.43]{ppgpeak.png}
%\caption{Original PPG signal (top), PPG signal with detected peaks (bottom)}
%\label{ppg_f}
%\end{figure}

\subsubsection{ECG Feature Extraction} A total of 22 features were extracted from the ECG signal based on Heart Rate (HR) and Heart Rate Variability (HRV). These features were calculated by applying algorithms in \cite{hrv_code} to extract time and frequency domain features from the ECG signal. The time-domain features include statistical features relating to HRV and the mean, maximum, minimum, and standard deviation of the heart rate. The frequency-domain features include the total power in the signal, the power in the very low frequency (VLF, 0.003 to 0.04 Hz), low frequency (LF, 0.04 to 0.15 Hz), and high frequency (HF, 0.15 to 0.40 Hz) bands, the ratio of the LF to HF bands, and the normalized LF and HF power.

\subsection{Feature Expansion} In the feature expansion unit, all of the 42 total features extracted from the GSR, Respiration, and ECG signals were expanded into 6 new features. These new features consist of the mean, median, standard deviation, minimum, maximum, and time-weighted average of the original features in groups of $n$ data points each. In order to focus on predicting transitions between stress levels, only the last $n$ data points for each driving section were expanded and passed to the next stage. To find an optimal value for $n$, four expanded feature sets are generated by performing feature expansion with $n =$ 2, 3, 4, and 5.

Each data point is also labeled low, medium, or high stress depending on the driving situation (Rest, Highway, or City, respectively), and the label and 252 features are packed into a data frame. At this point, the labels for each driving section were pushed back such that each data point is labeled with the upcoming driving situation. This results in the last data point for each drive being dropped because they cannot be used to predict a scenario. Because we are not interested in predicting the Rest situation, the data points which are labeled low stress are also dropped.

%In the next step, the training feature and the training labels are separated in to two different input lines to the Feature Selection Unit.

%\begin{figure}[ht]
%\centering
%\includegraphics[trim= 0cm 0cm 0cm 0cm, scale=0.35]{feature_freq.png}
%\caption{Selected Feature Frequencies}
%\label{feature_freq}
%\end{figure}
%
%\subsubsection{Feature Selection Unit} This unit performs feature selection, which is used mainly to reduce overfitting in the model by reducing the number of features. Feature selection is performed by training a random forest classifier with 100 estimators using the gini function and a maximum depth of 30 on the training data. The 10 features with the highest importance in the fitted classifier are selected as input to the model. This step was performed at each step of the leave-one-subject-out testing approach, a summary of selected features is available in Figure \ref{feature_freq}.

\section{Results}
\label{res}
In this section, we will discuss the performance of the fitted models on the test data. The model is trained by fitting a random forest classifier with 100 estimators using the Gini function and a maximum depth of 30 to the training set. The train and test sets are selected using the leave-one-subject-out testing approach, meaning that a drive is selected for the test set with the remaining drives composing the train set. After the model is evaluated, a different drive is selected for the test set such that each drive is selected once. This approach is repeated with each of the four expanded feature sets. The model is evaluated based on accuracy classifying the training and test data, and F1-score on the test data. The precision, recall, and F1-score of the model on the test data are also measured for the high and low-stress categories.

\begin{figure}[t]
\centering
\includegraphics[trim= 0cm 0cm 0cm 0cm, scale=0.60]{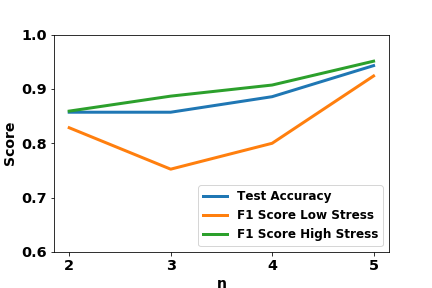}
\caption{Model Performance as $n$ Changes, where $n$ is the number of timesteps used in input data}
\label{perf_by_n}
\end{figure}

From Figure \ref{perf_by_n} it may be seen that the model has good performance for all tested values of n. Additionally, there appears to be a positive, linear relationship between $n$ and model performance. The clear exception to this relationship is the F1-score of the low-stress category for $n =$ 2, which is better than $n =$ 3 or 4. Table \ref{perf_by_n_detail} displays the low stress F1-score in more detail. It is clear that the low average performance is due to poor performance in classifying the low-stress situation in drive 10. This discrepancy appears to be the result of an unusual physiological reaction by the driver.

The best results were generated from the $n =$ 5 case.
%From Table \ref{perf} it may be seen that the model has excellent accuracy in classifying the training data. The table also shows that the worst case test accuracy and F1-score are acceptable, while the average for test accuracy and F1-score is 89\% and 0.88, respectively.
Table \ref{perf2} shows the average precision and recall of the model. The precision and recall for the low-stress scenario are 0.95 and 0.93, respectively, which indicates that the model both predicts the low-stress state effectively and has a low rate of incorrectly classifying low-stress as high-stress. The precision and recall for the high stress scenario are 0.96 and 0.95, respectively, which indicates that the model both predicts the high stress state effectively and has a low rate of incorrectly classifying high stress as low stress. The performance data indicates that our hypothesis has promise. A stress model $M$ trained on $x(t-1:t-n)$ has predicted $y(t)$ with good accuracy, as stated in our hypothesis.

\newcolumntype{g}{>{\columncolor{Gray}}c}
\newcolumntype{l}{>{\columncolor{LightCyan}}c}
\begin{table}[t]
\centering
\caption{Low Stress F1-Score as $n$ Changes}
\scalebox{1}{
\begin{tabular}{|g|l|g|l|g|l|g|l|}
\hline
 & & & & & & & \\
n$\backslash$Drive No. \ &\textbf{5} &\textbf{6} &\textbf{7} &\textbf{10} &\textbf{11} &\textbf{12} &\textbf{15} \\
 & & & & & & & \\
%\hline
\textbf{2} & 0.80  & 1.00 & 1.00 & 0.67 & 0.67 & 1.00 & 0.67 \\
 & & & & & & & \\
\textbf{3} & 0.80  & 1.00 & 1.00 & 0.00 & 1.00 & 0.80 & 0.67 \\
 & & & & & & & \\
\textbf{4} & 0.80  & 1.00 & 1.00 & 0.00 & 1.00 & 0.80 & 1.00 \\
 & & & & & & & \\
\textbf{5} & 0.80  & 1.00 & 1.00 & 0.67 & 1.00 & 1.00 & 1.00 \\
\hline
\end{tabular}
\label{perf_by_n_detail}
}
\end{table}

%\newcolumntype{g}{>{\columncolor{Gray}}c}
%\newcolumntype{l}{>{\columncolor{LightCyan}}c}
%\begin{table}[h]
%\centering
%\caption{Performance of the Predictive Model}
%\scalebox{1}{
%\begin{tabular}{|g|l|g|l|}
%\hline
% & & & \\
% &\textbf{Train Accuracy} &\textbf{Test Accuracy} &\textbf{F1-Score} \\
% & & & \\
%%\hline
%Best & 100\%  & 100\% & 1.0 \\
% & & & \\
%Worst & 100\% & 60\% & 0.58 \\
% & & & \\
%Average & 100\% & 89\% & 0.88 \\
%\hline
%\end{tabular}
%\label{perf}
%}
%\end{table}

\newcolumntype{g}{>{\columncolor{Gray}}c}
\newcolumntype{l}{>{\columncolor{LightCyan}}c}
\begin{table}[h]
\centering
\caption{Precision and Recall of the Predictive Model}
\scalebox{1}{
\begin{tabular}{|g|l|g|l|}
\hline
 & & & \\
 &\textbf{Precision} &\textbf{Recall} &\textbf{F1-Score} \\
 & & & \\
%\hline
Low Stress & 0.95  & 0.93 & 0.92 \\
 & & & \\
High Stress & 0.96 & 0.95 & 0.95 \\
 & & & \\
Weighted Average & 0.96 & 0.94 & 0.94 \\
\hline
\end{tabular}
\label{perf2}
}
\end{table}

\section{Discussions}
\label{discuss}
The physiological data used in this study is composed only of data from subjects as they completed a specific driving route through Boston. While the consistency of the data has the desirable property of revealing differences between individuals, it also makes it very difficult to create a generalized model. As a result, the current model may perform more poorly on data collected from a different driving route. Additionally, the ground truth used to train this model assumes that the road type is an accurate sole indicator of subjective stress. While this is a reasonable starting point, subjective stress could also be influenced by other drivers, weather conditions, or other occurrences which can vary independently of the road type.

\section{Future Research Direction}
\label{future_research}
Data from different driving routes would improve the stress prediction model by refining the assumptions that it makes about how subjective stress is influenced by driving conditions. Another method of improving the model would be using a better subjective stress indicator as the ground truth, which could allow the model to account for more than just the effect that the road type has on the driver. This indicator would likely need to have an improved sample rate to more closely match the rate at which subjective stress can change. Cortisol has been explored as an indicator of stress \cite{cortisol_stress} \cite{cortisol_stress2}, making it a potential candidate. With the aid of a newer type of sensor \cite{continuous_cortisol}, it can also be measured noninvasively in only a few seconds, making it an even more attractive candidate. Replacing the Random Forest Classifier with a deep neural network could also improve the model by removing the need for feature extraction and expansion. This would reduce the computational complexity of the model, thereby reducing the requirements for the edge device it operates on. A recurrent deep neural network has the added advantage of having the inherent ability to account for time dependency in input data, making it a good choice for this task.

\section{Conclusion}
\label{conc}
In this work, we have presented a stress prediction model which can predict the stress level of a subject up to one minute in advance. The model uses GSR, Respiratory, and ECG data taken while the subject is driving. The model then predicts whether the stress level of the subject will be high using $n$ time steps of data prior to the period of interest. Performance data indicates an approximately linear increase in performance with increasing $n$. The best performance of the model was at $n =$ 5, where the model has an average test accuracy of 94\%. This indicates that the model has good performance and could be expanded to include other driving situations.

%\section{Acknowledgments}

% I don't think I have anything to put here %This work was supported by the Kentucky Science and Engineering Foundation under Grant KSEF-3528-RDE-019.
\bibliographystyle{IEEEtran}
\bibliography{references}

\begin{thebibliography}{10}
\providecommand{\url}[1]{#1}
\csname url@samestyle\endcsname
\providecommand{\newblock}{\relax}
\providecommand{\bibinfo}[2]{#2}
\providecommand{\BIBentrySTDinterwordspacing}{\spaceskip=0pt\relax}
\providecommand{\BIBentryALTinterwordstretchfactor}{4}
\providecommand{\BIBentryALTinterwordspacing}{\spaceskip=\fontdimen2\font plus
\BIBentryALTinterwordstretchfactor\fontdimen3\font minus
  \fontdimen4\font\relax}
\providecommand{\BIBforeignlanguage}[2]{{%
\expandafter\ifx\csname l@#1\endcsname\relax
\typeout{** WARNING: IEEEtran.bst: No hyphenation pattern has been}%
\typeout{** loaded for the language `#1'. Using the pattern for}%
\typeout{** the default language instead.}%
\else
\language=\csname l@#1\endcsname
\fi
#2}}
\providecommand{\BIBdecl}{\relax}
\BIBdecl

\bibitem{drive_stress}
\BIBentryALTinterwordspacing
G.~Matthews, L.~Dorn, T.~W. Hoyes, D.~R. Davies, A.~I. Glendon, and R.~G.
  Taylor, ``Driver stress and performance on a driving simulator,'' \emph{Human
  Factors}, vol.~40, no.~1, pp. 136--149, 1998, pMID: 9579108. [Online].
  Available: \url{https://doi.org/10.1518/001872098779480569}
\BIBentrySTDinterwordspacing

\bibitem{drive_stress2}
A.~Queyam and M.~Singh, ``Correlation between physiological parameters of
  automobile drivers and traffic conditions,'' \emph{International Journal of
  Electronics Engineering}, vol.~5, pp. 6--12, 12 2013.

\bibitem{youngjun_stress_detect}
Y.~Cho, S.~J. Julier, and N.~Bianchi-Berthouze,
  ``\BIBforeignlanguage{eng}{Instant stress: Detection of perceived mental
  stress through smartphone photoplethysmography and thermal imaging},''
  \emph{\BIBforeignlanguage{eng}{JMIR mental health}}, vol.~6, no.~4, pp.
  e10\,140--e10\,140, 2019.

\bibitem{martino_stress_detect}
F.~D. Martino and F.~Delmastro, ``\BIBforeignlanguage{eng}{High-resolution
  physiological stress prediction models based on ensemble learning and
  recurrent neural networks},'' in \emph{\BIBforeignlanguage{eng}{2020 IEEE
  Symposium on Computers and Communications (ISCC)}}.\hskip 1em plus 0.5em
  minus 0.4em\relax IEEE, 2020, pp. 1--6.

\bibitem{stress_sensors}
H.~J. Baek, H.~B. Lee, J.~S. Kim, J.~M. Choi, K.~K. Kim, and K.~S. Park,
  ``\BIBforeignlanguage{eng}{Nonintrusive biological signal monitoring in a car
  to evaluate a driver's stress and health state},''
  \emph{\BIBforeignlanguage{eng}{Telemedicine journal and e-health}}, vol.~15,
  no.~2, pp. 182--189, 2009.

\bibitem{stress_intervention}
H.~{Thapliyal}, V.~{Khalus}, and C.~{Labrado}, ``Stress detection and
  management: A survey of wearable smart health devices,'' \emph{IEEE Consumer
  Electronics Magazine}, vol.~6, no.~4, pp. 64--69, 2017.

\bibitem{healey_stress_detect}
J.~Healey and R.~Picard, ``\BIBforeignlanguage{eng}{Detecting stress during
  real-world driving tasks using physiological sensors},''
  \emph{\BIBforeignlanguage{eng}{IEEE transactions on intelligent
  transportation systems}}, vol.~6, no.~2, pp. 156--166, 2005.

\bibitem{db_times}
A.~Akbaş, ``Evaluation of the physiological data indicating the dynamic stress
  level of drivers,'' 2011.

\bibitem{gsr_code}
``Scipy,'' \url{https://www.scipy.org/}, accessed: 2021-1-25.

\bibitem{resp_code}
``Biosppy,'' \url{https://biosppy.readthedocs.io/}, accessed: 2021-1-25.

\bibitem{hrv_code}
``hrv-analysis,'' \url{https://pypi.org/project/hrv-analysis/}, accessed:
  2021-1-25.

\bibitem{cortisol_stress}
S.~Betti, R.~M. Lova, E.~Rovini, G.~Acerbi, L.~Santarelli, M.~Cabiati, S.~D.
  Ry, and F.~Cavallo, ``\BIBforeignlanguage{eng}{Evaluation of an integrated
  system of wearable physiological sensors for stress monitoring in working
  environments by using biological markers},''
  \emph{\BIBforeignlanguage{eng}{IEEE transactions on biomedical engineering}},
  vol.~65, no.~8, pp. 1748--1758, 2018.

\bibitem{cortisol_stress2}
\BIBentryALTinterwordspacing
R.~K. Nath, H.~Thapliyal, and A.~Caban-Holt, ``Machine learning based stress
  monitoring in older adults using wearable sensors and cortisol as stress
  biomarker,'' \emph{Journal of Signal Processing Systems}, Jan 2021. [Online].
  Available: \url{https://doi.org/10.1007/s11265-020-01611-5}
\BIBentrySTDinterwordspacing

\bibitem{continuous_cortisol}
\BIBentryALTinterwordspacing
O.~Parlak, S.~T. Keene, A.~Marais, V.~F. Curto, and A.~Salleo, ``Molecularly
  selective nanoporous membrane-based wearable organic electrochemical device
  for noninvasive cortisol sensing,'' \emph{Science Advances}, vol.~4, no.~7,
  2018. [Online]. Available:
  \url{https://advances.sciencemag.org/content/4/7/eaar2904}
\BIBentrySTDinterwordspacing

\end{thebibliography}

\end{document}